\documentclass[10pt,twocolumn,letterpaper]{article}

\usepackage{cvpr}              % To produce the CAMERA-READY version
\definecolor{cvprblue}{rgb}{0.21,0.49,0.74}
\usepackage[pagebackref,breaklinks,colorlinks,allcolors=cvprblue]{hyperref}

\def\paperID{} % *** Enter the Paper ID here
\def\confName{CVPR}
\def\confYear{2026}

\title{LangFlash: Feed-forward 3D Language Gaussian Splatting from Sparse Unposed Images}

\author{
Yilong Liu$^{1,3}$ \quad
Wanhua Li$^{1,2,\ast}$ \quad
Chen Zhu-Tian$^{4}$ \quad
Hanspeter Pfister$^{1}$\\
$^{1}$Harvard University \quad
$^{2}$Nanyang Technological University\\
$^{3}$Tsinghua University \quad
$^{4}$University of Minnesota - Twin Cities\\[2mm]
{\footnotesize
\begin{tabular}{c}
\texttt{liuyilon22@mails.tsinghua.edu.cn, wanhua.li@ntu.edu.sg, ztchen@umn.edu, pfister@seas.harvard.edu}
\end{tabular}}
}

\usepackage{multirow}
\usepackage{algorithm}     
\usepackage{algpseudocode}
\usepackage{placeins}  
\usepackage{caption}   
\begin{document}

\twocolumn[{%
\renewcommand\twocolumn[1][]{#1}%
\maketitle
\begin{center}
    \centering
    \captionsetup{type=figure}
    \includegraphics[width=1\textwidth]{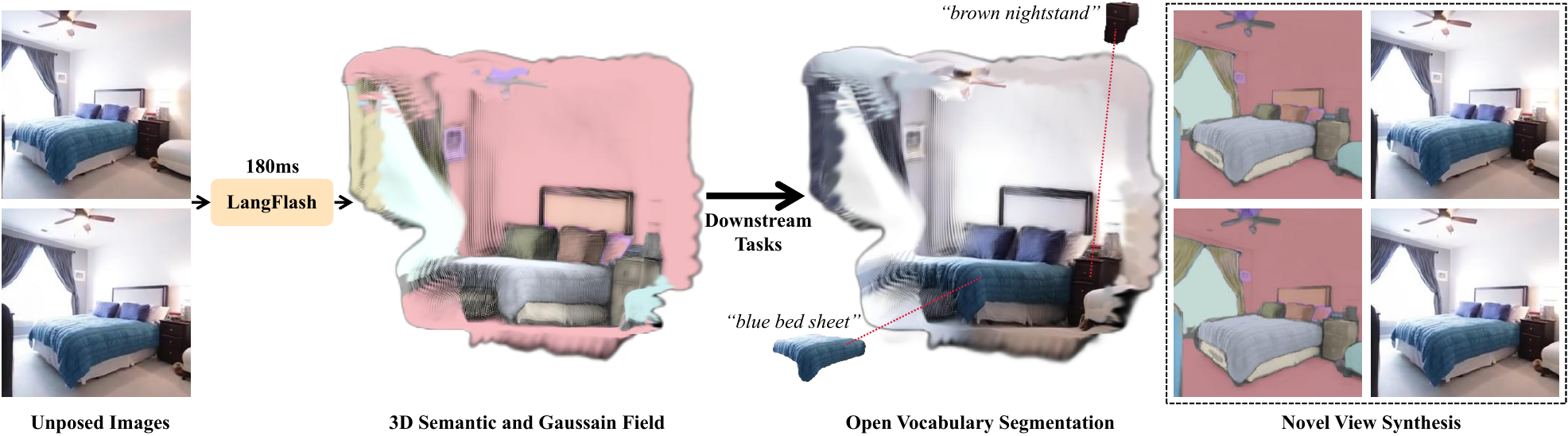}
    % \vspace{-22pt}
    \captionof{figure}{LangFlash reconstructs 3D semantic Gaussian fields directly from sparse unposed multi-view images in a single forward pass. Given unposed input images, LangFlash predicts geometry and semantics within 180 ms, producing a unified 3D Semantic and Gaussian Field that supports multiple downstream tasks, including open-vocabulary 3D segmentation and novel view synthesis. 
    }
    % \vspace{-3pt}
    \label{fig:teaser}
% \vspace{-6pt}
\end{center}%
}]
{\let\thefootnote\relax\footnotetext{{$^{\ast}$ Corresponding author}}}

\maketitle
\begin{abstract}
We present LangFlash, a feed-forward framework for 3D Language Gaussian Splatting that reconstructs 3D scenes parameterized by Gaussian primitives enriched with language-aligned semantic features from sparse unposed multi-view images. Unlike optimization-based 3D methods, LangFlash directly predicts the geometry and semantics in a single forward pass, enabling low-latency 3D reconstruction and language-consistent scene understanding. To support large-scale training, we enriched the RealEstate10k dataset with coherent and dense semantic information for 3D semantic supervision. Furthermore, we propose a sparse semantic encoding scheme that combines a global semantic dictionary with locally
varying per-primitive weights, preserving high-level linguistic information, while reducing representation complexity. Experimental results show that LangFlash achieves superior novel view synthesis and semantic consistency compared with previous methods. This study establishes a new paradigm for pose-free, language-grounded 3D scene reconstruction, advancing generalizable 3D vision and multimodal scene understanding. Code is available at \href{https://liylo.github.io/langflash.github.io/}{https://liylo.github.io/langflash.github.io/}.
\end{abstract}    

\section{Introduction}
3D scene reconstruction~\cite{tancik2022block,fan2025momentum} is a fundamental problem in computer vision. Recent advances have significantly improved the realism of applications in embodied AI and robotics, while also enabling novel view synthesis to enhance environmental perception and understanding \cite{Duan2022, ZhuRobotics2024}.

Despite progress on both optimization-based \cite{Mildenhall2020, Kerbl2023} and feed-forward \cite{Charatan2024, Chen2024, Smart2024, Zhang2025, Jiang2025} methods, current approaches remain restricted to geometry and appearance, without incorporating semantics. However, semantics are essential for high-level reasoning and decision-making because they provide structured representations of the environment. Therefore, a unified framework that jointly reconstructs the geometry, appearance, and semantics in 3D is desirable. Moreover, existing semantic models \cite{liu2023grounding,li2022,zang2022open,ma2022open,xu2023multi} typically operate on 2D observations and cannot embed semantics into 3D representations, limiting novel view inference and comprehensive scene understanding.

Earlier attempts have incorporated semantics into NeRF \cite{Zhi2021, Liu2023sem, Kerr2023,garfield2024}, but the inefficiency of implicit representations results in poor performance under sparse-view inputs. More recent efforts have extended 3DGS to semantic fields \cite{qin2024langsplat,li2025langsplatv2,zhou2024feature3dgs,wu2024opengaussian,ye2024gaussiangrouping,li20254d}, yet these still rely on per-scene optimization and lack cross-scene generalization. Other attempts \cite{chen2025,hu2025,li2025semanticsplat} directly project fine-grained pixel-aligned language features onto Gaussian points predicted by feed-forward models, allowing the reconstruction of the semantic field in a relatively faster manner. However, constructing such fine-grained features requires time-consuming segmentation and tracking models as preprocessing steps, making them unsuitable for real-time reconstruction. 

The latest LSM framework \cite{Fan2024} moves toward the feed-forward estimation of geometry, appearance, and semantics, offering notable efficiency over optimization-based approaches. However, its strategy of directly downsampling and regressing high-dimensional language features fundamentally limits the quality of semantic reconstruction. High-dimensional embeddings contain fine-grained semantic cues, and aggressive downsampling inevitably discards the key semantic information and contextual relations required for accurate 3D grounding. The resulting features also exhibit cross-view inconsistencies, and regressing these noisy, high-dimensional vectors imposes a heavy training burden, which slows convergence and leads to unstable semantic predictions. Furthermore, storing and rendering such high-dimensional features during inference is computationally expensive. These limitations underscore the need for a more compact, structured, and semantically aligned framework that preserves linguistic meaning while remaining efficient for feed-forward 3D reconstruction.

To address these limitations, we propose LangFlash, a novel generalizable framework that synthesizes a unified 3D representation from arbitrary sparse images for both high-fidelity rendering and dense open-vocabulary semantic understanding. The proposed model is powered by a large dataset, RealEstate10k (RE10k) \cite{realestate10k2023} (over 10M frames), with semantic information that we have labeled. By leveraging the \emph{Semantic Grouping} and \emph{Language Feature Aggregation} modules, our approach effectively integrates information from multi-view RGB images and semantic maps, yielding globally consistent representations. LangFlash predicts unified 3D Gaussian primitives that are enriched with open-vocabulary semantic features. These Gaussian representations can be seamlessly rendered with low latency  to synthesize novel views,  bypassing the need for per-scene optimization.

To balance the semantic expressiveness and computational efficiency in feature prediction, we introduced a sparse feature-encoding scheme. Specifically, we first construct a high-dimensional global dictionary from the input image that captures a wide range of semantic information. Instead of directly predicting high-dimensional feature vectors, the model estimates only sparse weights corresponding to the entries in the global dictionary. This design significantly reduces the computational and storage overheads while preserving the expressive power of the semantic features.

Our contributions are summarized as follows.
\begin{itemize}
    \item We enrich the RE10k \cite{realestate10k2023} dataset with coherent semantic information, where every pixel in every frame of each video is associated with a high-quality, temporally consistent dimensional semantic feature.
    \item We introduce LangFlash, a novel feed-forward architecture that unifies 3D reconstruction and semantic understanding. Instead of directly regressing language features, it predicts a global semantic dictionary with locally varying, per-primitive weights. 
    \item Experimental results show that our method not only achieves superior performance in both novel view synthesis and 3D open-vocabulary segmentation, but also delivers an inference speed of only 180 ms per scene.
\end{itemize}
\begin{figure*}[t] 
    \centering
    \includegraphics[width=0.7\linewidth]{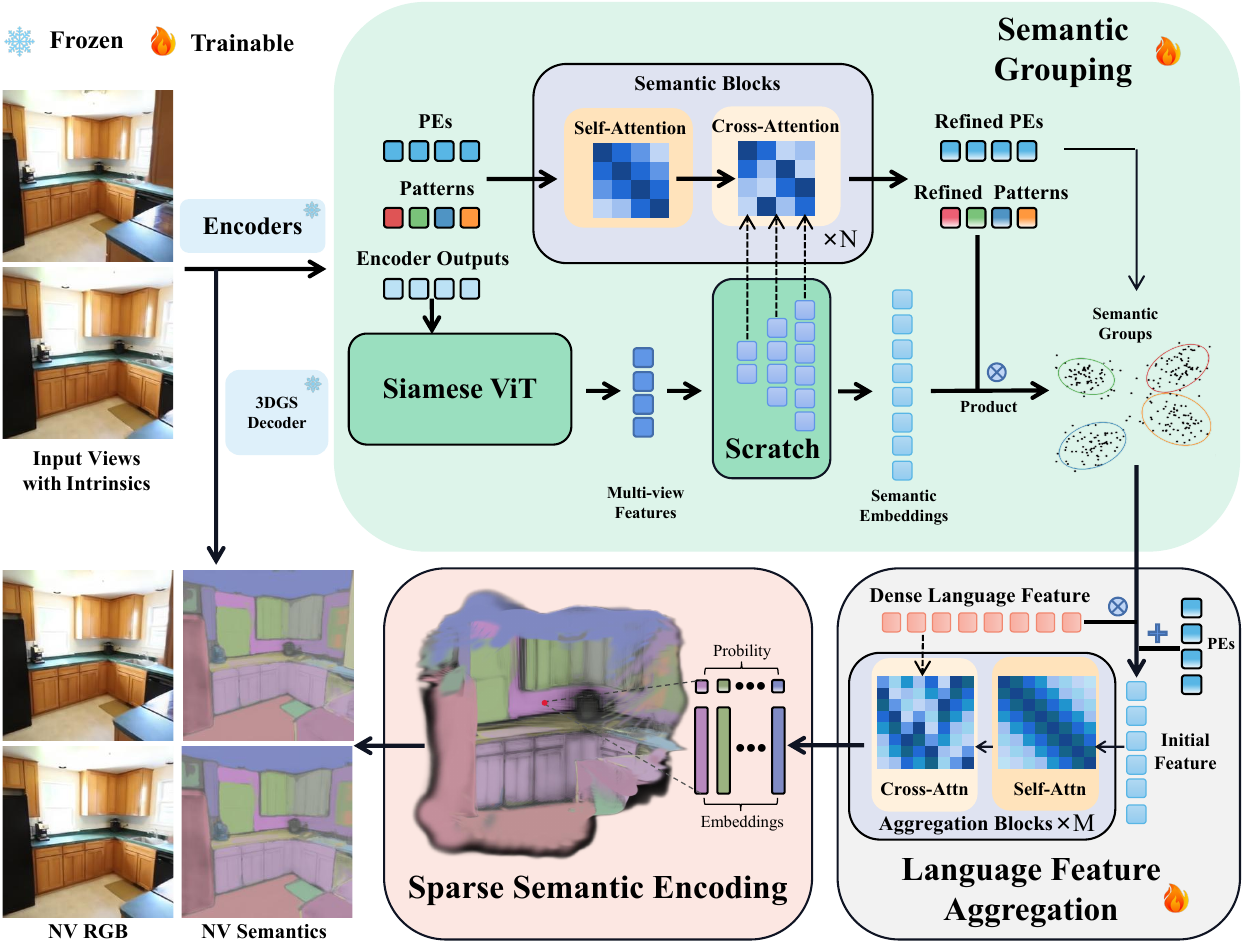}
    \caption{The overall architecture of LangFlash is illustrated as follows. Features extracted by the shared image encoders are first passed to a pretrained 3D Gaussian decoder to reconstruct scene geometry. In parallel, learnable positional embeddings, learnable patterns, and encoder outputs are processed by the semantic grouping module. The resulting groups, together with the refined positional embeddings, are then fed into the language feature aggregation stage, where coarse dense language features from a pretrained CLIP-like model are further refined to produce the final embeddings for each selected group. Outputs from all preceding modules are finally integrated to form the complete 3D semantic representation.}
    \label{fig:pipeline}
\end{figure*}
\section{Related Works}

\textbf{Feed-Forward Reconstruction.}
Classical methods such as NeRF \cite{Mildenhall2020} and 3D Gaussian Splatting \cite{Kerbl2023} produce high-quality novel views by optimizing a scene-specific representation at test time, which is computationally expensive and unsuitable for real-time use. In contrast, feed-forward approaches learn priors from large datasets and perform one-shot or few-shot inference without per-scene optimization; these methods directly predict 3D representations (e.g., point maps, point clouds, or Gaussian primitives) from sparse, possibly unposed images, enabling fast and generalizable reconstruction \cite{Charatan2024, Chen2024, Smart2024, Zhang2025, Jiang2025}.
Recent studies have further eliminated the need for known camera poses by predicting the geometry and appearance in a canonical frame from uncalibrated inputs. In particular, transformer-based 3D reconstruction priors (e.g., DUSt3R~\cite{Wang2024}, MASt3R~\cite{Leroy2024}, and VGGT~\cite{wang2025}) cast matching and reconstruction as 3D pointmap regression and enable pose-free or pose-agnostic reconstruction. Building on these geometric priors, Gaussian-based feed-forward methods such as Splatt3R~\cite{Smart2024} and other zero-/few-shot splatting approaches directly infer splat parameters from uncalibrated images for fast rendering \cite{ ye2025,wang2024freesplat,jiang2025anysplat}. Overall, feed-forward models demonstrate impressive inference speed and cross-scene generalization, making them attractive for embodied agents and interactive systems \cite{Charatan2024, Chen2024}.

\noindent \textbf{Semantic-aware 3D Representations.}
Embedding semantics into 3D representations has been explored in both implicit (NeRF-based) and explicit (e.g., Gaussian splatting) families. Early implicit efforts extended NeRF to include semantic outputs supervised by 2D labels, enabling novel-view semantic synthesis but inheriting the computational cost of per-scene optimization \cite{Zhi2021, Liu2023sem, Kerr2023,garfield2024}.
Explicit 3D Gaussian approaches have recently been extended to include semantic components or to distill 2D semantic models into 3D primitives, enabling open-vocabulary or dense semantic 3D representations \cite{qin2024langsplat,zhou2024feature3dgs,wu2024opengaussian,ye2024gaussiangrouping}. However, many of these studies still rely on per-scene optimization. More recently, generalizable feed-forward models \cite{Fan2024} jointly predict geometry, appearance, and semantics from unposed inputs in a single forward pass, moving the field toward pose-free, semantic-aware, and real-time 3D understanding. 
% Our LangFlash follows this trend: it operates in a pose-free, feed-forward regime using 3D Gaussian primitives while tightly integrating open-vocabulary semantic features into the representation, aiming to balance efficiency and reconstruction/segmentation quality.
Our LangFlash follows this direction by operating in a pose-free, feed-forward framework built upon 3D Gaussian primitives while tightly integrating open-vocabulary semantic features into the representation to achieve a balance between efficiency and reconstruction accuracy.

\section{Method}

\subsection{Feed-Forward Semantic Gaussian Splatting}
\label{subsec:Feed-Forward Semantic Gaussian Splatting}
\paragraph{Overview.}
Our model accepts unposed input images together with corresponding camera intrinsics and outputs an efficient semantic Gaussian field, which will be described in \ref{subsec:Sparse Semantic Encoding}. We first build on a pretrained NoPospalt \cite{ye2025} network and freeze it. The features from its encoder are then fed into a \emph{Semantic Grouping} module to produce semantic groups. Subsequently, the semantic groups, together with language features extracted by a pretrained vision-language model \cite{li2022}, are processed by a \emph{Language Feature Aggregation} module to obtain fine-grained features for each semantic group, which are then used to form the scene's semantic Gaussian representation.

\paragraph{Semantic Grouping.}
The Semantic Grouping module takes the encoder's multi-view outputs as input. We employ a Siamese ViT-based decoder to \cite{weinzaepfel2023} produce multi-scale outputs for every view, and a scratch head refines each view's multi-scale outputs \cite{ranftl2021} to produce multi-view, multi-scale dense semantic maps. We concatenate these refined outputs along the view dimension to form a scene-level, multi-scale dense feature field \(F(\mathbf{x})\) defined at each spatial location, \(\mathbf{x}\). Finally, we add 1D sinusoidal positional encodings \cite{vaswani2017attention} to the scene's multi-scale outputs.

Following the DETR-like \cite{carion2020detr,zhang2022dino,li2023maskdino,cheng2021maskformer,liu2022dabdetr} designs, we maintain \(N\) learnable pattern vectors (queries) \(\{q_n\}_{n=1}^N\) and positional embeddings \(\{r_n\}_{n=1}^N\). We add positional embeddings to the queries in all subsequent layers. The processed queries undergo self-attention for refinement. Subsequently, cross-attention between the refined queries and dense features provides the queries with knowledge of the 3D structure of the scene.
Each query eventually passes through an MLP head to predict its group existence probability \(p_n(\mathbf{x})\) at spatial location \(\mathbf{x}\) in the scene. To obtain a compact and variable-length representation, we filter out inactive queries using the threshold \(p_n(\mathbf{x}) > \tau_{\text{exist}}\), resulting in \(K\) valid semantic groups (\(K \le N\)). For these surviving \(K\) groups, we apply a softmax operation across the group dimension to obtain normalized probability maps \(w_k(\mathbf{x})\) for \(k = 1, \dots, K\). For a Gaussian primitive \(i\) located at \(\mathbf{x}_i\), this continuous probability directly yields its semantic weight \(w_{ik} = w_k(\mathbf{x}_i)\), which will be explained in Section~\ref{subsec:Sparse Semantic Encoding}.

\paragraph{Language Feature Aggregation.}
We utilize a pretrained vision-language model \cite{li2022} to extract a downsampled language feature map \(L(\mathbf{x})\), which is bilinearly interpolated back to the input resolution.

Using the derived semantic weights, the initial feature for each valid group \(k\) is aggregated as
\begin{equation}
\mathbf{d}_k^{(0)} = \frac{\sum_{\mathbf{x}} w_k(\mathbf{x})\,L(\mathbf{x})}{\sum_{\mathbf{x}} w_k(\mathbf{x}) + \epsilon}.
\label{eq:initial-aggregate}
\end{equation}

Subsequently, our aggregation block updates this initial group feature using multiple transformer blocks. Within each block, we conduct cross-attention between the group features and the language feature map \(L(\mathbf{x})\) to obtain the final refined semantic dictionary atom, \(\mathbf{d}_k\). From the Semantic Grouping and Language Feature Aggregation modules, we obtain the per-primitive weights \(\mathbf{w}_i\) and the global dictionary \(\mathcal{D} = \{\mathbf{d}_k\}_{k=1}^K\), providing sufficient elements to reconstruct the 3D semantic Gaussian field detailed in Section~\ref{subsec:Sparse Semantic Encoding}.

% \paragraph{Loss}

\paragraph{Semantic Grouping Loss.}  
The predicted groups are matched to the ground-truth groups using Hungarian matching \cite{carion2020detr}. For matched groups, we apply a focal loss \(L_{\text{focal}}\) \cite{lin2017focal}  and dice loss \(L_{\text{dice}}\) \cite{milletari2016v}. The unmatched groups are supervised using an existence loss \(L_{\text{exist}}\) \cite{carion2020detr}.

We also apply dense supervision from a SAM \cite{Kirillov2023} encoder output \(S_{\text{SAM}}(\mathbf{x})\) using:
\begin{equation}
L_{\text{MSE}} = \left\|F(\mathbf{x}) - S_{\text{SAM}}(\mathbf{x})\right\|_2^2.
\label{eq:mse-loss}
\end{equation}

Total Semantic Grouping loss:
\begin{equation}
L_{\text{SG}} = \lambda_{\text{foc}} L_{\text{focal}} + \lambda_{\text{dice}} L_{\text{dice}} + \lambda_{\text{exist}} L_{\text{exist}} + \lambda_{\text{mse}} L_{\text{MSE}}.
\label{eq:sg-total}
\end{equation}

\paragraph{Language Feature Aggregation Loss.} 
For the Language Feature Aggregation (LFA) module, we supervise the refined feature with three objectives: consistency with the input feature, alignment to the ground-truth language feature, and, when text labels are available, additional alignment to the text embedding.
Moreover, if text labels exist, we modulate the GT alignment weight using the cosine similarity between the GT feature (its derivation will be discussed in \ref{subsec:Continuous Semantic Label Collection}, donated as $g_n^{\mathrm{gt}}$) and the text embedding (denoted as $t_n$), so that only positive-correlated pairs contribute.

The input consistency loss is defined as follows:
\begin{equation}
L_{\text{in}} = \frac{1}{N} \sum_{n=1}^{N} \left(1 - \cos\big(\mathbf{d}_n,  \mathbf{d}_n^{(0)}\big)\right).
\label{eq:lin}
\end{equation}

The GT alignment loss is:
\begin{equation}
L_{\text{gt}} = \frac{1}{N} \sum_{n=1}^{N} p_n \left(1 - \cos\big(\mathbf{d}_n, g_n^{\mathrm{gt}}\big)\right),
\label{eq:lgt}
\end{equation}
where the adaptive weight \(p_n\) is defined as:
\begin{equation}
w_n =
\begin{cases}
\max\{0,\,\cos(g_n^{\mathrm{gt}}, t_n)\}, & \text{if text label } t_n \text{ is available},\\[4pt]
1, & \text{otherwise}.
\end{cases}
\label{eq:wn}
\end{equation}

When text labels are available, we further apply text alignment loss as follows:
\begin{equation}
L_{\text{text}} = \frac{1}{N} \sum_{n=1}^{N} \left(1 - \cos\big(\mathbf{d}_n, t_n\big)\right).
\label{eq:ltext}
\end{equation}

The total LFA loss combines all terms, where the text-related term is activated only when text labels are provided:
\begin{equation}
L_{\text{LFA}} = \lambda_{\text{in}} L_{\text{in}}
+ \lambda_{\text{gt}} L_{\text{gt}}
+ \mathbb{I}_{\{\text{text}\}} \lambda_{\text{text}} L_{\text{text}}.
\label{eq:llfa}
\end{equation}
Here, \(\mathbb{I}_{\{\text{text}\}}\) is an indicator function that equals 1 when the text labels are available and 0 otherwise.

\paragraph{Training Schedule}
We first train the Semantic Grouping module using \(L_{\text{SG}}\) while freezing geometry. Then, we freeze them all and train the Language Feature Aggregation module using \(L_{\text{LFA}}\). This schedule stabilizes the convergence and improves the consistency of the multiview.

\subsection{Sparse Semantic Encoding}
\label{subsec:Sparse Semantic Encoding}
We represent the semantics of a scene using sparse encoding that combines a global semantic dictionary with locally varying per-primitive weights. As described in \ref{subsec:Feed-Forward Semantic Gaussian Splatting}, given sparse input images, we first reconstruct the Gaussian field and predict the corresponding semantic groups. Next, we extract semantic features $\mathcal{F}(\mathbf{x})$ using a pretrained language feature extractor \cite{li2022}. Based on these semantic groups, we then pool and refine the features to obtain the global semantic dictionary
\begin{equation}
\mathcal{D} = \{ \mathbf{d}_k \in \mathbb{R}^C \}_{k=1}^K,
\label{eq:dictionary}
\end{equation}
where $K$ is the variable dictionary length and $C$ is the feature dimension of each dictionary atom. 

Each Gaussian primitive $i$ in the scene is assigned a weight vector $\mathbf{w}_i = [w_{i1}, \dots, w_{iK}]^\top$ over the dictionary, so that the semantic feature associated with primitive $i$ is the linear combination
\begin{equation}
\mathbf{f}_i \;=\; \sum_{k=1}^K w_{ik}\,\mathbf{d}_k.
\label{eq:primitive-feature}
\end{equation}

We adopted the standard volumetric compositing used in Gaussian splatting. Let the primitives be arranged along a ray. Denote by $\alpha_i$ the opacity (alpha) of primitive $i$ and by $T_i$ the front-surface transmittance up to $i$:
\begin{equation}
T_i \;=\; \prod_{j < i} \bigl(1 - \alpha_j\bigr).
\label{eq:transmittance}
\end{equation}
Replacing the usual RGB color $\mathbf{c}_i$ with the semantic feature $\mathbf{f}_i$, the rendered pixel feature $\mathbf{F}_p$ at pixel (or ray) $p$ is
\begin{align}
\mathbf{F}_p
&= \sum_{i} T_i\,\alpha_i\,\mathbf{f}_i
= \sum_{i} T_i\,\alpha_i \left(\sum_{k=1}^K w_{ik}\,\mathbf{d}_k\right)
\label{eq:feature-first}\\
&= \sum_{k=1}^K \left(\sum_{i} T_i\,\alpha_i\,w_{ik} \right)\mathbf{d}_k
= \sum_{k=1}^K W_{k}(p)\,\mathbf{d}_k,
\label{eq:weight-first}
\end{align}
where we define the \emph{rendered weight map} for dictionary atom $k$ as
\begin{equation}
W_k(p) \;=\; \sum_{i} T_i\,\alpha_i\,w_{ik}.
\label{eq:weight-map}
\end{equation}

Equations \eqref{eq:feature-first}--\eqref{eq:weight-first} show that computing the per-primitive features $\mathbf{f}_i$ and then compositing yields exactly the same result as first composing scalar weights into per-atom weight maps $W_k(p)$ and then linearly combining dictionary atoms. Because $K \ll C$ and the dictionary atoms $\mathbf{d}_k$ are globally shared, it is far more efficient to (i) render the scalar weight maps $\{W_k(p)\}_{k=1}^K$ and then (ii) form the final feature via a small matrix multiplication
\begin{equation}
\mathbf{F}_p \;=\; \mathbf{D} \mathbf{W}(p),
\label{eq:matrix-form}
\end{equation}
where
\begin{equation}
\mathbf{D} = [\mathbf{d}_1, \dots, \mathbf{d}_K] \in \mathbb{R}^{C \times K}, 
\label{eq:D-def}
\end{equation}
\begin{equation}
\mathbf{W}(p) = [W_1(p), \dots, W_K(p)]^\top \in \mathbb{R}^K.
\label{eq:W-def}
\end{equation}
This factorized encoding provides a practical balance between expressiveness and efficiency. The global dictionary captures reusable semantic concepts shared across scenes, whereas the per-primitive weights preserve local variations and spatial specificity. Because the representation separates appearance-independent semantic atoms from their spatial assignments, it can be rendered efficiently while remaining compatible with the Gaussian splatting pipeline.

\subsection{Continuous Semantic Label Collection}
\label{subsec:Continuous Semantic Label Collection}
Our continuous semantic label collection aims to establish 3D semantic paired data by associating each pixel $P_i$ with a rich semantic feature $\mathbf{F}_i \in \mathbb{R}^d$.

\begin{algorithm}[t]
\caption{Continuous Semantic Label Collection}
\label{alg:collection}
\begin{algorithmic}[1]

\Require Video frames $\{I_t\}_{t=1}^T$, SAMv2, SAM, CLIP
\Ensure Pixel–feature pairs $\{(P_i, F_i)\}_{i=1}^N$

\For{each frame $I_t$}
    \State Generate candidate masks using SAMv2 and SAM
    \State Apply post-NMS filtering to remove redundant masks
    \State Identify new objects using propagated masks
    \If{first frame or new objects detected}
        \State Propagate masks using SAMv2 predictor
    \EndIf
    \State Store object masks $\{M_{t,k}\}$
\EndFor

\For{each pixel $P_i$ in mask $M_{t,k}$}
    \State $F_i \gets \text{CLIP}(P_i)$
    \State Save $(P_i, F_i)$
\EndFor

\State \Return $\{(P_i, F_i)\}_{i=1}^N$ on RE10K

\end{algorithmic}
\end{algorithm}

As outlined in Alg.~\ref{alg:collection}, we employ SAMv2~\cite{Ravi2024} and SAM~\cite{Kirillov2023} for continuous object-level segmentation. 
We then use CLIP~\cite{Radford2021} for semantic extraction. 
This results in a collection of pixel-feature pairs 
$\{(P_i, F_i)\}_{i=1}^N$ across the RE10k  dataset \cite{realestate10k2023}, providing a solid foundation for training. As we obtain segmentation groups, pixels within the nth group share the same gt language feature, which can also be regarded as $g_n^\mathbf{gt}$  as described in \ref{subsec:Feed-Forward Semantic Gaussian Splatting}.

The proposed collection pipeline is intended to reduce the semantic ambiguity commonly found in open-world video data. By combining mask generation, propagation, and post-filtering, we obtain more stable object-level regions over time. These regions provide a consistent basis for extracting continuous semantic features, which are essential for learning dense 3D semantic correspondences.

\section{Experiments}

We conduct experiments on ScanNet~\cite{dai2017scannet} and  3D-OVS~\cite{liu2023weakly} to validate the effectiveness of LangFlash.
Section~\ref{sec:impl} describes the implementation details of the proposed method.
Section~\ref{sec:semantic3d} presents the quantitative results of semantic 3D reconstruction and open-vocabulary segmentation.
Section~\ref{sec:ablation} provides ablation studies that analyze the impact of each design component.

\begin{figure*}[t] 
    \centering
    \includegraphics[width=1.0\linewidth]{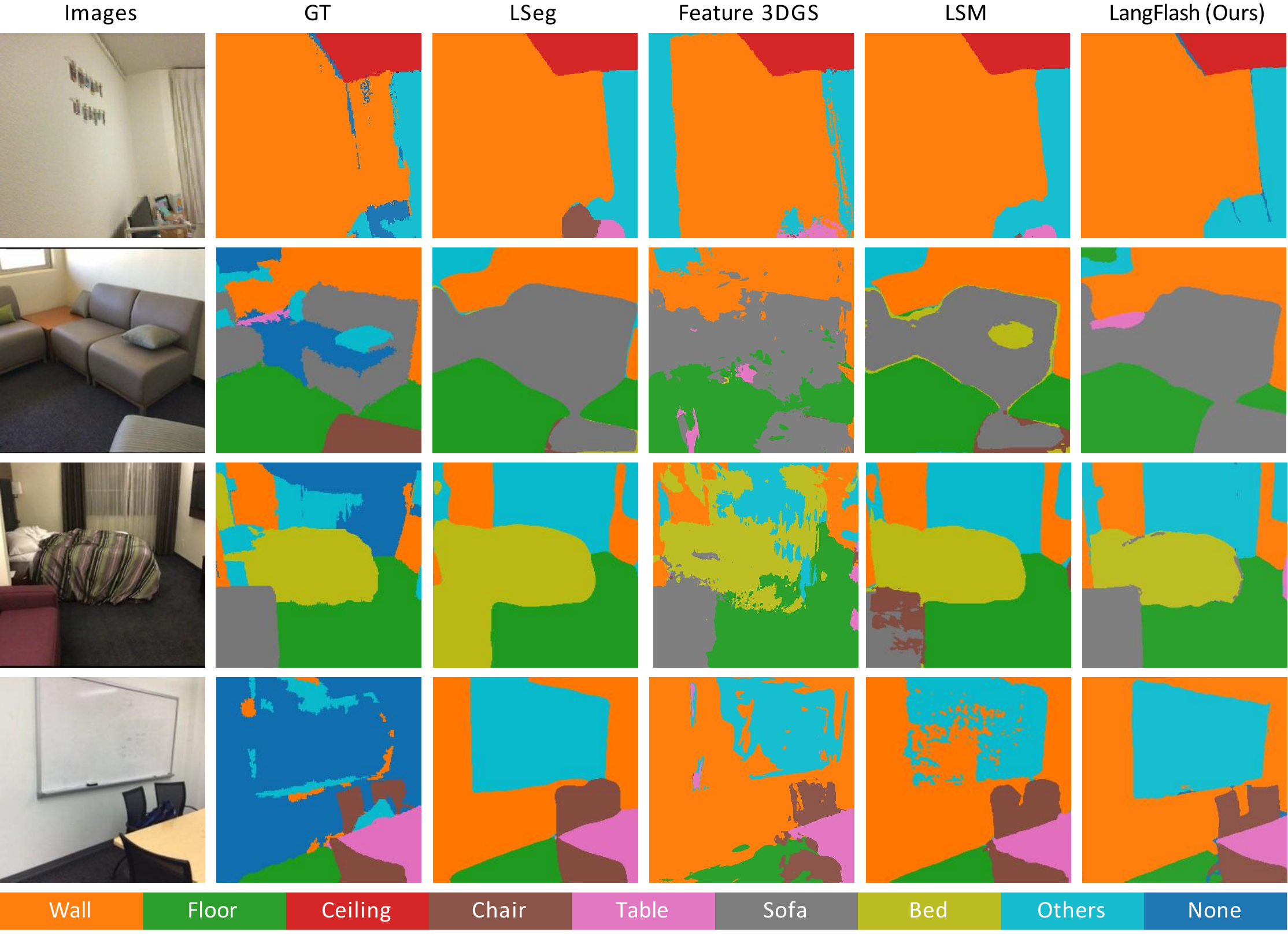}
    \caption{Language-based 3D Segmentation Comparison on ScanNet \cite{dai2017scannet}. We visualize the segmentation results
across four unseen scenes and observe that our method performs comparably to previous methods.}
    \label{fig:vis}
\end{figure*}
\begin{table*}[t]
\centering
\caption{Quantitative Comparison in 3D Tasks on ScanNet \cite{dai2017scannet}.
We report the novel view synthesis and open-vocabulary segmentation accuracy. 
Our method eliminates the need for any preprocessing in 3D tasks while achieving performance comparable to other baselines that rely on SfM to obtain camera poses. }
\label{tab:main}
\resizebox{0.99\textwidth}{!}{
\begin{tabular}{l|cc|cc|cc|ccc}
\toprule
& \multicolumn{2}{c|}{Reconstruction Time$\downarrow$}   
& \multicolumn{2}{c|}{Source View}                                                                                                   
& \multicolumn{2}{c|}{Target View (Segmentation)}  
& \multicolumn{3}{c}{Target View (Rendering)} \\ 
\cline{2-10}
& SfM         
& Per-Scene                
& mIoU $\uparrow$       
& Acc.$\uparrow$        
& mIoU $\uparrow$       
& Acc.$\uparrow$        
& PSNR $\uparrow$ 
& SSIM $\uparrow$ 
& LPIPS $\downarrow$  \\ 
\midrule
LSeg \cite{li2022}~         
& N/A & N/A 
& 0.5278  & 0.7654                
& 0.5281 &  0.7612              
& - & - & - 
\\
NeRF-DFF \cite{kobayashi2022distilledfeaturefields}~  
& 20.52s & 1min2s
& 0.4540 & 0.7173 
& 0.4037 & 0.6755 
& 19.86 & 0.6650 & 0.3629 
\\
Feature-3DGS \cite{zhou2024feature3dgs}~ 
& 20.52s & 18min36s                 
& 0.4453 & 0.7276 
& 0.4223 & 0.7174 
& 24.49 & 0.8132 & 0.2293 
\\
pixelSplat \cite{Charatan2024}    
& 20.52s & 0.064s
& - & -                                                            
& - & -                 
& \textbf{24.89} & \textbf{0.8392} & \textbf{0.1641}             
\\
LSM~ \cite{Fan2024}        
& \multicolumn{2}{c|}{\textbf{0.108s} }                
& 0.5034 & 0.7740 
& 0.5078 & 0.7686 
& 24.39 & 0.8072 & 0.2506 
\\
\midrule
Ours (zero-shot)~         
& \multicolumn{2}{c|}{0.180s}                 
& 0.6217 & 0.7878 
& 0.6265 & 0.7861 
&  24.80 &  0.7906 & 0.2072
\\
Ours~         
& \multicolumn{2}{c|}{0.180s}                 
& \textbf{0.7344} & \textbf{0.8746} 
&  \textbf{0.7416} &  \textbf{0.8718}
&  24.80 &  0.7906 & 0.2072
\\
\bottomrule
\end{tabular}}
\end{table*}

\subsection{Implementation Details}
\label{sec:impl}

We initialize the geometry prediction layers using NoPoSplat \cite{ye2025}, and optimize the entire system end-to-end with the loss function described above. For a fair comparison with the LSM \cite{Fan2024}, we train two versions of the model separately on ScanNet \cite{dai2017scannet} and RE10k \cite{realestate10k2023}. Specifically, the model is trained for 10k steps on ScanNet \cite{dai2017scannet} and 50k steps on RE10k \cite{realestate10k2023}. We use AdamW \cite{loshchilov2017decoupled} as the optimizer with a base learning rate of 2e-4 for all experiments.
For ScanNet \cite{dai2017scannet}, evaluation is conducted on 40 unseen scenes, following the same protocol as LSM \cite{Fan2024}. In addition to novel-view synthesis, we evaluate the model on 3D language-based semantic segmentation, measuring its ability to align open-vocabulary queries with consistent 3D representations. For 3D-OVS \cite{liu2023weakly}, evaluation is conducted on 5 complex scenes, focusing exclusively on 3D language-based semantic segmentation.

\begin{figure*}[t] 
    \centering
    \includegraphics[width=1.0\linewidth]{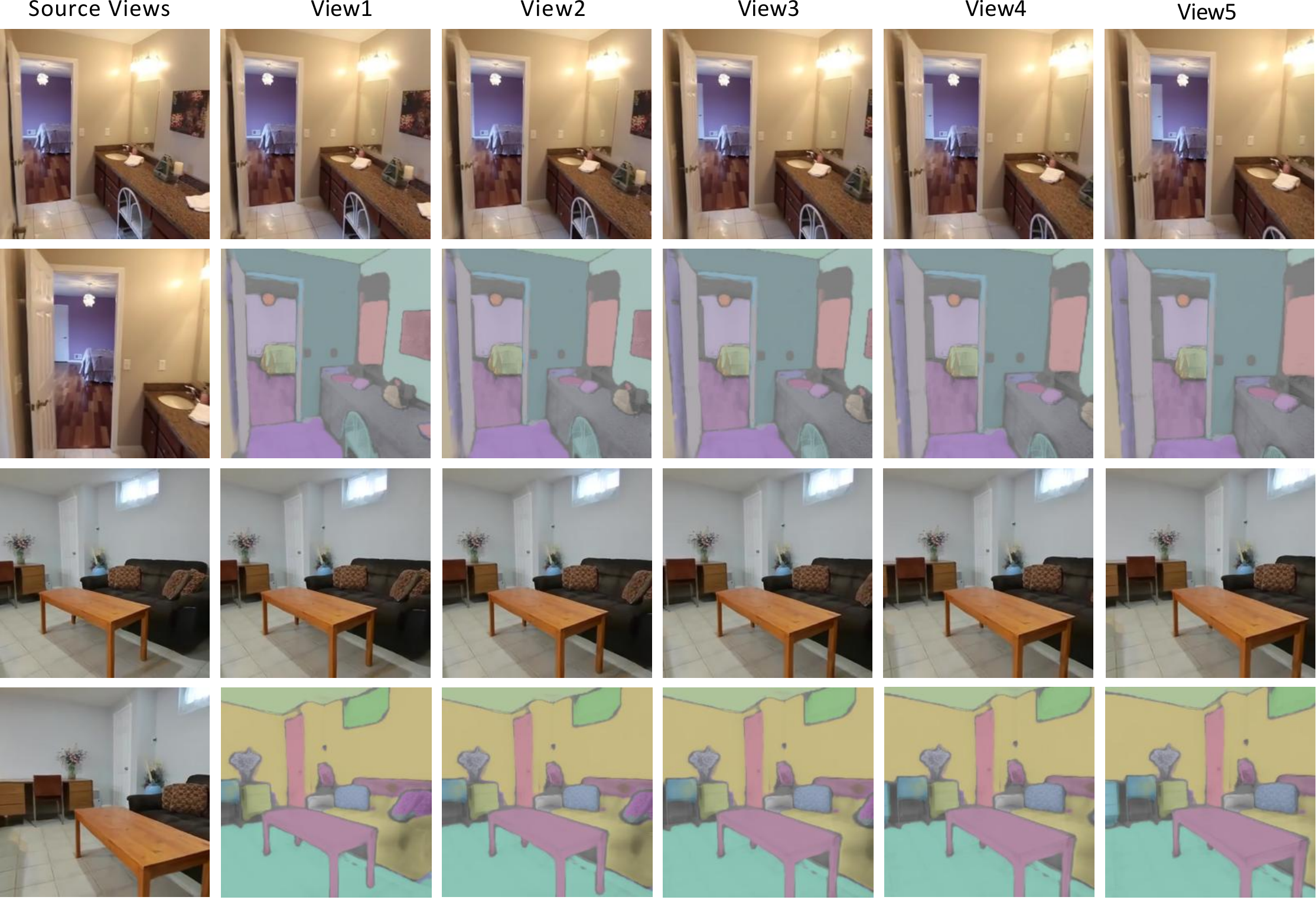}
    \caption{Qualitative results on Re10K.
We further visualize both the segmentation and novel-view synthesis results, which demonstrate that our method reconstructs 3D-consistent semantic Gaussian fields with high fidelity.}
    \label{fig:vis2}
    \vspace{-10pt}
\end{figure*}

\subsection{Semantic 3D Reconstruction}
\label{sec:semantic3d}

In Table~\ref{tab:main}, our method achieves the highest accuracy on both the source and target views, demonstrating that the proposed 3D semantic Gaussian field delivers strong open-vocabulary segmentation while maintaining cross-view consistency. Compared with prior 3D approaches, LangFlash shows substantial gains in target-view accuracy, indicating a more discriminative and robust geometry–language representation. Notably, the model trained only on RE10k \cite{realestate10k2023} (``Ours (zero-shot)''), generalizes well to ScanNet \cite{dai2017scannet}, confirming strong cross-dataset transfer. Moreover, LangFlash reconstructs each scene in 0.180s without SfM preprocessing, achieving an effective balance between efficiency and accuracy for real-world robotics and embodied AI. The qualitative results (Figures~\ref{fig:vis} and \ref{fig:vis2}) show semantically coherent structures across views, with consistent boundaries and accurate localization, even in cluttered scenes. The semantic field maintains stable region assignments under large viewpoint changes and preserves object-level continuity in the absence of artifacts. On the 3D-OVS dataset \cite{liu2023weakly} (Table~\ref{table:3dovs}), our zero-shot model outperforms both 2D and 3D methods, including LERF \cite{Kerr2023}, demonstrating that LangFlash captures object-level concepts and larger regions reliably. This robustness persists under uneven category distributions and varying lighting and geometries. Failure cases mainly occur for very small or thin objects, reflective surfaces, or fine-grained ambiguous categories but do not affect the overall performance. Overall, LangFlash provides a strong, generalizable, and efficient solution for open-vocabulary 3D semantic segmentation using its 3D-consistent Gaussian representation and robust language–geometry alignment. 

\begin{table}[t]
  \caption{Quantitative comparisons of 3D semantic segmentation on the 3D-OVS dataset \cite{liu2023weakly}. We report the mIoU scores (\%).}
  \renewcommand\tabcolsep{3pt}
  \centering
  \begin{tabular}{lcccccc}
    \toprule
    Method & \textit{bed}   & \textit{bench} & \textit{room}  & \textit{sofa}  & \textit{lawn}  & overall   \\
    \midrule
    LSeg~\cite{li2022}      & 56.0           & 6.0            & 19.2           & 4.5            & 17.5           & 20.6           \\
    ODISE~\cite{Xu2023}     & 52.6           & 24.1           & 52.5           & \textbf{48.3}           & 39.8           & 43.5           \\
    \midrule
    FFD~\cite{kobayashi2022distilledfeaturefields}       & 56.6           & 6.1            & 25.1           & 3.7            & 42.9           & 26.9           \\
    LERF~\cite{Kerr2023}      & \textbf{73.5 }          & 53.2           & 46.6           & 27             & 73.7         & 54.8            \\
    \midrule
    Ours (zero shot)   & 67.8 & \textbf{83.2} & \textbf{64.7} & 33.3 & \textbf{75.7} & \textbf{64.9} \\ 
    \bottomrule
  \end{tabular}
  \label{table:3dovs}
  % \vspace{-10pt}
\end{table}

\subsection{Ablation Studies}
\label{sec:ablation}
We perform ablation experiments to validate the effectiveness of our design choices. All experiments evaluate both language-driven segmentation and novel-view synthesis, with quantitative segmentation results reported in Table~\ref{tab:ablation_quant}. It summarizes the results: PointTransformer-based \cite{zhao2021point,wu2022point,wu2024ptv3} segmentation (PT, Exp. ~[1]) is evaluated using ground-truth masks and Hungarian matching; SG with ground-truth masks shows clear improvement (Exp.~[2]); SG with averaged LSeg \cite{li2022} language features yields coarser results (Exp.~[3]); and SG with our proposed LFA achieves the best performance (Exp.~[4]). These experiments isolate each component under controlled conditions and demonstrate how different design choices affect the overall performance.

\noindent\textbf{Semantic Grouping.} We implement a variant of our segmentation module using PointTransformer blocks \cite{zhao2021point,wu2022point,wu2024ptv3} (denoted as 'PT'). However, PT relies on highly accurate 3D point coordinates, which are difficult to obtain in a feedforward Gaussian model owing to inherent regression noise. Consequently, the PT-based design is sensitive to small spatial perturbations, which is evident even when evaluated using ground-truth masks. The PT variant shows a clear performance drop compared to our Semantic Grouping (SG) module (Exp.~[1] vs.~[2] in Table~\ref{tab:ablation_quant}), indicating that PT struggles to maintain stable associations under coordinate noise. In contrast, the SG operates reliably in this setting, maintaining a coherent grouping even with imperfect geometry, thereby demonstrating its robustness to geometric variations in feed-forward Gaussian pipelines.

\begin{table}[ht]
\centering
\caption{Quantitative results of the ablation study for different segmentation implementations and feature aggregation designs.}
\label{tab:ablation_quant}
\renewcommand\tabcolsep{10pt}
\begin{tabular}{c|l|cc}
\toprule
\multirow{1}{*}{Exp ID} & Model & mIoU$\uparrow$ & Acc.$\uparrow$ \\
\midrule
{[}1] & PT w. gt masks & 0.6536 & 0.8277 \\
{[}2] & SG w. gt masks & \textbf{0.7601} & \textbf{0.9214} \\
{[}3]& SG w. avg feature & 0.7273 & 0.8601 \\
{[}4]  & SG w. LFA & 0.7416 & 0.8718 \\
\bottomrule
\end{tabular}
% \vspace{-8pt}
\end{table}

\noindent\textbf{Language Feature Aggregation.} Language–visual models, such as LSeg \cite{li2022} tend to exhibit local alignment between image and text features. Because of this locality, a naive strategy, such as the simple averaging of per-point language features, can only produce coarse and noisy associations, especially in regions where multiple semantic concepts are spatially intertwined or where language cues vary gradually across the scene. In contrast, our proposed module enables the model to retain finer distinctions while preserving contextual relationships. As shown in Table~\ref{tab:ablation_quant}, replacing the LFA with simple averaging (Exp. ~[3]) leads to a weaker performance. While using our LFA (Exp. ~[4]) yields noticeably stronger results, reflecting that our aggregation strategy can more effectively emphasize discriminative cues and suppress ambiguous cues.  

\noindent\textbf{Dictionary length K.}We adopt $K=128$ as the default setting. To analyze the impact of this hyperparameter, we compare different values of $K$ on ScanNet \cite{dai2017scannet} using equal training time. Table~\ref{tab:different_k_scannet} validates that our selected value provides a favorable trade-off between performance and speed.

\begin{table}[ht]
  \centering
  \caption{Results of Different $K$ on ScanNet}
  \label{tab:different_k_scannet}
  \resizebox{\columnwidth}{!}{
    \begin{tabular}{lccccc}
      \toprule
      Metric & $K=64$ & $K=96$ & $K=128$ & $K=196$ & $K=256$ \\
      \midrule
      mIoU $\uparrow$ & 0.6928 & 0.6577 & 0.7144 & 0.7165 & \textbf{0.7228} \\
      Acc. $\uparrow$  & 0.7953 & 0.5131 & \textbf{0.8547} & 0.8348 & 0.8439 \\
      \bottomrule
    \end{tabular}
  }
\end{table}

\noindent\textbf{Runtime analysis.}  
As shown in Table~\ref{tab:time_breakdown}, the main computational cost arises from the LFA (72 ms), followed by the SG (38 ms). In contrast, LSeg \cite{li2022} and the Gaussian decoder are relatively lightweight. The end2end runtime is 180 ms per scene (2 input views), demonstrating its efficiency.
\begin{table}[ht]
  \centering
  \caption{Computation Time for Different Components}
  \label{tab:time_breakdown}
  \resizebox{\columnwidth}{!}{
    \begin{tabular}{lcccccc}
      \toprule
      & LSeg & Shared Encoder  & SG & LFA &  Gaussian Decoder & All \\
      \midrule
      Time (ms) & 6 & 32  & 38 & 72 & 32 & 180 \\
      \bottomrule
    \end{tabular}
  }
\end{table}

\section{Conclusion}

In this study, we present \textbf{LangFlash}, a generalizable and low-latency framework for 3D reconstruction and semantic understanding from sparse views. LangFlash integrates open-vocabulary language features with efficient 3D Gaussian Splatting, predicting geometry, appearance, and semantics in a single forward pass, thereby eliminating the need for per-scene optimization. To support large-scale training and ensure semantic consistency, we enrich the RE10k dataset with temporally coherent semantic annotations and design a novel pipeline comprising \emph{Semantic Grouping} and \emph{Language Feature Aggregation} modules, which extract globally consistent group-wise representations from multi-view RGB images and language features. Our dictionary-based \emph{Sparse Scene Encoding} scheme further compresses the representation, enabling more efficient training while preserving the semantic fidelity and reconstruction accuracy. Experiments show that LangFlash not only supports instant rendering but also delivers substantially improved open-vocabulary semantic reconstruction, demonstrating its strong efficiency and generalization.
% Notably, LangFlash builds on an off-the-shelf reconstruction model and inherits its limitations, with degraded performance under extremely wide fields of view or limited view overlap. Future studies may improve the results by adopting a stronger backbone.
% Nevertheless, LangFlash has potential areas for improvement: semantic label quality is influenced by language feature extraction and joint learning of geometry and semantics—rather than stage-wise freezing—may further improve representation and performance.  Future work could focus on exploring end-to-end deployment and task-driven fine-tuning of LangFlash in real-time applications such as mobile robotics and augmented reality. 

% \input{rebuttal}

% \clearpage
\section*{Acknowledgements} 
This work was supported in part by the NIH grant R01HD104969 and the NTU Nanyang Assistant Professorship Startup Grant 025661-00012. This research was also supported in part by Google.org and the Google Cloud Research Credits Program through the Gemma Academic Program.

{
    \small
    \bibliographystyle{ieeenat_fullname}
    \bibliography{main}
}

\clearpage
\clearpage
\setcounter{page}{1}
\maketitlesupplementary

% --- SECTION 1: Qualitative visualizations (double-column image at top) ---
\section{RE10k Qualitative visualizations}

\begin{figure*}[!t]
  \centering
  \includegraphics[width=\textwidth]{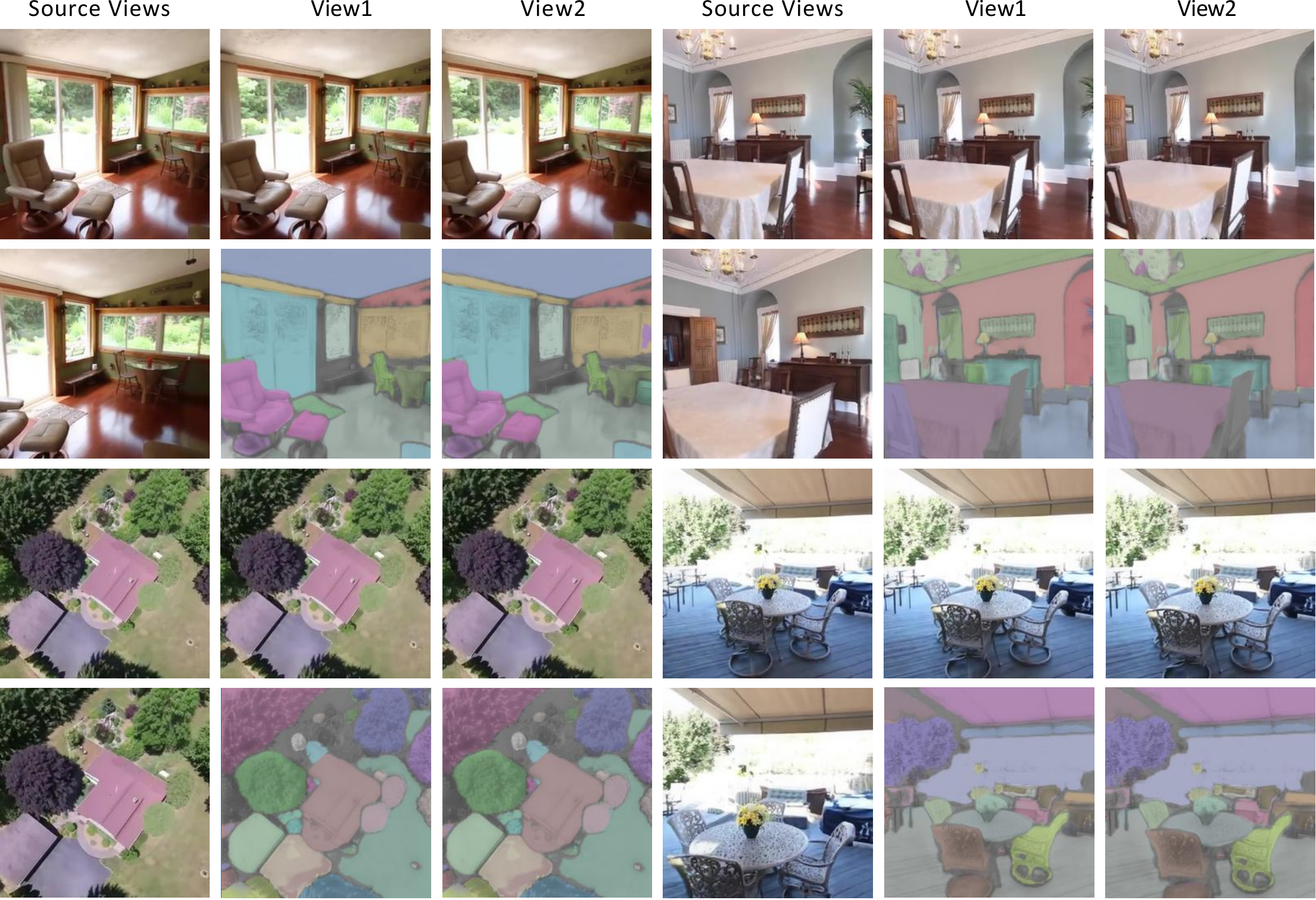}
  \caption{Additional qualitative results on RE10k. We visualize both the semantic and novel-view synthesis results.}
  \label{fig:vis3}
\end{figure*}

The visual results shown in \cref{fig:vis3} provide a qualitative overview of our performance on the RE10k dataset. These examples were selected to emphasize the characteristic challenges in the dataset: numerous small and overlapping object instances, wide lighting variation, and strong viewpoint changes that stress both 2D segmentation and multi-view rendering. These qualitative results complement the numerical summary.

% --- SECTION 2: Dataset statistics ---
\section{RE10k Dataset statistics}

\begin{table}[!t]
  \centering
  \caption{Statistics of the processed RE10k dataset.}
  \label{tab:re10k_stats}
  \begin{tabular}{l r}
    \toprule
    Metric & Value \\
    \midrule
    Total frames & $\sim$6M \\
    Number of scenes & $\sim$10k \\
    Avg. masks per image & 36 \\
    Avg. mask coverage per image (\%) & 83.5 \\
    Processing time (single NVIDIA A100) & 30 days \\
    \bottomrule
  \end{tabular}
\end{table}

The table above (\cref{tab:re10k_stats}) summarizes the primary corpus-level statistics of the processed RE10k split used in this study. In total, we retained approximately six million frames across roughly ten thousand scenes; on average, each image contained dozens of instance masks, with mask pixels covering the majority of the image area. The reported processing time corresponds to running the full pipeline (mask extraction, per-frame cleanup, and multi-view consolidation) on a single NVIDIA A100; in practice, the pipeline is embarrassingly parallel, and the wall-clock time can be reduced by distributed execution.

% --- SECTION 3: 3D semantic segmentation results ---
\section{RE10k 3D semantic segmentation}

\begin{table*}[!t]
  \centering
  \caption{3D semantic segmentation on RE10k (mIoU, \%). We assign scene names (absent in the original dataset) and provide their corresponding original identifiers.}
  \label{table:re10k}
  \renewcommand\tabcolsep{3pt}
  \begin{tabular}{lcccccc}
    \toprule
    Method & \textit{Bedroom(5aca)} & \textit{Aisle(bc95)} & \textit{Living room (6558)} & \textit{Study room(89ea)} & \textit{Pool(cd74)} & Overall \\
    \midrule
    LSeg & 29.60 & 18.64 & 19.12 & 45.84 & 14.48 & 25.53 \\
    LSM  & 24.39 & 11.67 & 23.33 & 42.95 & 24.77 & 25.42\\
    \midrule
    Ours & \textbf{34.33} & \textbf{22.37} & \textbf{34.97} & \textbf{57.44} & \textbf{37.99} & \textbf{37.42} \\
    \bottomrule
  \end{tabular}
\end{table*}

In addition to \ref{fig:vis2}, we annotated five previously unseen scenes and report the per-scene mIoU as well as the average overall score in \cref{table:re10k}. The baseline methods (LSeg and LSM) struggled on several scenes, whereas our method achieved substantially higher per-scene and overall mIoU, indicating more consistent cross-view semantic aggregation. These results validate both the quality of the processed dataset and the effectiveness of our proposed approach for 3D semantic segmentation of large-scale, real-world indoor footage.

\end{document}